\documentclass{article}

    \PassOptionsToPackage{numbers, compress}{natbib}

    \usepackage[preprint]{neurips_2024}

\usepackage[utf8]{inputenc} 
\usepackage[T1]{fontenc}    
\usepackage{hyperref}       
\usepackage{url}            
\usepackage{booktabs}       
\usepackage{amsfonts}       
\usepackage{nicefrac}       
\usepackage{microtype}      
\usepackage{xcolor}         

\usepackage{multirow}
\usepackage{array}
\usepackage{graphbox}
\usepackage{color}
\usepackage{graphicx}
\usepackage{comment}
\usepackage{amsmath,amssymb} 
\usepackage{soul}
\usepackage{float}
\usepackage{pifont}
\usepackage{wrapfig}
\usepackage{makecell}

\definecolor{linegreen}{RGB}{84, 130, 53}
\definecolor{lineorange}{RGB}{197, 90, 17}

\newcolumntype{?}[1]{!{\vrule width #1}}

\newcommand*{\ShowNotes}{}
\ifdefined\ShowNotes
  \newcommand{\colornote}[3]{{\color{#1}\bf{#2: #3}\normalfont}}
\else
  \newcommand{\colornote}[3]{}
\fi

\title{Image-aware Evaluation of Generated Medical Reports}

\author{Gefen Dawidowicz$^*$ \hspace{0.1in}
Elad Hirsch$^*$ \hspace{0.1in}
Ayellet Tal \\
Technion -- Israel Institute of Technology
}

\begin{document}

\maketitle

\def\thefootnote{*}\footnotetext{These authors contributed equally to this work}\def\thefootnote{\arabic{footnote}}

\vspace{-0.2in}
\begin{abstract}
The paper proposes a novel evaluation metric for automatic medical report generation from X-ray images, $VLScore$. 
It aims to overcome the limitations of existing evaluation methods, which either focus solely on textual similarities, ignoring clinical aspects, or concentrate only on a single clinical aspect, the pathology, neglecting all other factors. 
The key idea of our metric is to measure the similarity between radiology reports while considering the corresponding image. 
We demonstrate the benefit of our metric through evaluation on a dataset where radiologists marked errors in pairs of reports, showing notable alignment with radiologists' judgments. 
In addition, we provide a new dataset for evaluating metrics. 
This dataset includes well-designed perturbations that distinguish between significant modifications (e.g., removal of a diagnosis) and insignificant ones. 
It highlights the weaknesses in current evaluation metrics and provides a clear framework for analysis.
\end{abstract}

\section{Introduction}

Medical reports are crucial for summarizing diagnoses seen in X-ray images.
These reports facilitate communication among healthcare professionals, ensuring continuity of care and informed decision-making. 
The process of generating these reports requires expertise and time, typically involving a trained radiologist who writes them from scratch.
Automatic generation of medical reports therefore has the potential to enhance patient care and streamline administrative processes, ultimately improving healthcare delivery. 
Consequently, numerous research efforts have been directed toward this goal~\cite{chen-acl-2021-r2gencmn,chen-etal-2020-generating,jing2017automatic,li2019knowledge,tanida2023interactive,wang2022cross}.
These medical report generation models aim to produce coherent and accurate reports based on input images, with a current focus on chest radiology reports.

However, evaluating the generated reports is a challenge in its own right~\cite{yu2023evaluating}. 
This evaluation is essential to determine whether a new technique truly advances the state of the art.
Currently, two families of metrics are being used, language metrics and clinical metrics.
Derived from language generation tasks, such as neural machine translation or image captioning, generated medical reports are evaluated  with natural language generation (NLG) metrics.
In our context, the similarity between the generated report (candidate) to the ground-truth report (reference) associated with the input image is measured.
To evaluate some of the medical entities that are expressed in the report, clinical efficacy (CE) is measured by applying a pre-trained classification network on the ground-truth and output reports to assess their agreement on pre-defined pathologies~\cite{irvin2019chexpert,smit2020combining}.
These metrics suffer from drawbacks, which the current paper attempts to solve, as described hereafter.

The traditional NLG metrics, namely BLEU~\cite{papineni-etal-2002-bleu}, METEOR~\cite{banerjee-lavie-2005-meteor}, ROUGE~\cite{lin-2004-rouge}, and CIDEr~\cite{vedantam2015cider}, are n-gram-based and measure the token (or sequence of tokens) overlap between the candidate and reference texts.  
More recent metrics, such as BERTScore~\cite{zhang2019bertscore}, rely on deep textual feature similarity, making them more robust to variations in exact tokens and instead utilizing contextual embeddings. 
However, these metrics fail to capture several critical nuances in the context of medical reports. 
In particular, while the reports are typically lengthy and structured, following a systematic review method, they contain mostly normal findings and general references. The crucial findings may occupy a small portion of the report, sometimes comprising only one or two words (e.g., "large pleural effusion" vs. "no pleural effusions").
As these findings may be brief, they may not significantly impact the scores, despite their important information. 
Furthermore, radiologists frequently include in the report information that cannot be extracted from an image, such as the patient's history, findings from other sources, or details about treatment. 
These metrics penalize for such missing information.

CE metrics are oblivious to these issues and focus solely on assessing the presence or absence of certain pathologies in the reports.
However, due to the difficulty of comprehensively labeling the data, the available pathology classes are limited both in diversity and in detail. 
Currently, the number of pathologies used for evaluation is only $14$, neglecting possible findings beyond these pathologies.
Furthermore, only the existence or absence of a pathology is taken into account, while specifics such as locations (e.g., "left" vs. "right") and severity (e.g., "small" vs. "mild") are completely disregarded.
These limitations constrain the metrics' ability to capture overall clinical correctness.

Finally, an additional major drawback in the current situation in this domain is that results are evaluated separately on these two types of metrics.
This raises evaluation ambiguity: what does it signify if one model demonstrates improved scores in one type of metric but worse scores in the other?
For example, the result of \cite{li2023dynamic} is better than~\cite{wang2023metransformer} in terms of CE, but worse in terms of NLG metrics on MIMIC-CXR dataset~\cite{johnson2019mimic}. 
Which generative model shall one use?

Indeed, it was demonstrated in~\cite{yu2023evaluating} that these metrics exhibit a low correlation with radiologist judgment.
In this study, pairs of reference and candidate reports were evaluated by several NLG and CE metrics and compared to radiologist evaluations. 
To improve the correlation, it was proposed  to utilize  
a graph representation of clinical entities and relations~\cite{jain2021radgraph}, and to use it to compute an additional metric, 
by measuring the similarity in both the entities and their relationships between a reference report and a generated report.
However, this graph structure is not unique, as evidenced by the variability between manual annotators, which can potentially lead to inaccurate metrics. 
Additionally, normal findings may or may not appear in a report, and their presence or absence can influence the result, although both scenarios are valid.

\begin{figure}[t]
  \begin{center}
    \includegraphics[width=1\linewidth]{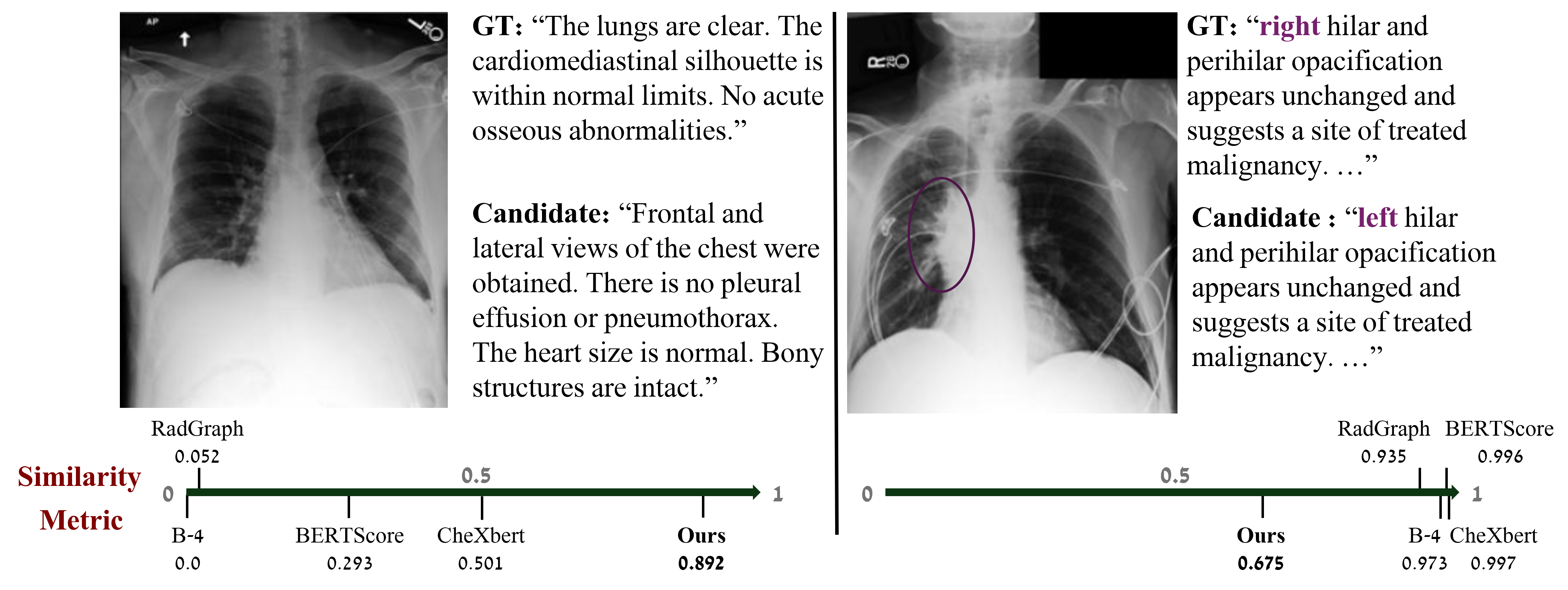}\\
  \end{center}
  \hspace{1in} (a) Same findings \hspace{1.5in} (b) Different location
\caption{
{\bf Vision-language based evaluation of generated reports.}
Our proposed evaluation metric for a reference report (GT) and a candidate report takes into account the associated image, thereby overcoming several drawbacks of common metrics that only measure the distance between two texts. 
Consequently, when comparing two semantically very similar reports that suit an image (a), our metric suggests a high degree of similarity, whereas other metrics are unaware of these similarities and therefore give low scores due to textual differences.
In contrast, when comparing two reports that are textually similar but differ by a single word that impacts the location of a finding (b), common metrics still provide a very high score, while our metric penalizes for that error.
}
\label{fig:teaser}
\end{figure}

We propose a novel metric to measure the correspondence between two reports.

The key idea is that the reports are linked through an associated image, and this image should be taken into account. 
Hence, we assess both the textual similarity between the reports  and the extent to which they convey shared clinical information.
By utilizing the visual modality, we address several weaknesses inherent in existing approaches discussed above. 
These include the separation to two distinct types of metrics, the overemphasis of general long phrases compared to important concise phrases in NLG metrics, the consideration of information that cannot be inferred from the image, and the limited diversity and detail captured by CE metrics.
Figure~\ref{fig:teaser} demonstrates the benefits of our proposed metric. 
When two reports are semantically very similar but significantly different in terms of text (Figure~\ref{fig:teaser}a), common metrics assign very low scores, whereas our metric manages to capture the high similarity between the reports.
In contrast, when two reports are highly similar textually except for a single word that holds significant anatomical importance (e.g., left vs. right), as in Figure~\ref{fig:teaser}b, common metrics provide a high score, while our metric is more sensitive to this critical change.

Specifically, our approach is based on the idea that measuring similarity between reports in a joint visual-textual space expresses aspects of similarity and distinctiveness, which cannot be achieved in a single modality.
This can be explained by the fact that reports conveying equivalent relevant information, which can be extracted from a given image, are expected to be positioned closely together within this space.
Furthermore, the learned space ensures that a broad spectrum of pathologies and observations are accommodated, rather than being limited to pre-defined ones.
Hence, we compute the proximity of $3$ components in this space -- the image, the ground-truth report and the generated report.
We note that a joint embedding space of images and reports has garnered interest in recent years within the field of representation learning in the medical domain, facilitating tasks such as retrieval and classification~\cite{boecking2022making,dawidowicz2023limitr,huang2021gloria,wang2022medclip,zhang2022contrastive}. 
This is the first time to propose that measuring similarities within this space provides a robust framework for evaluating how accurately a generated report describes an image.
Additionally, we define a new similarity score in this shared space.

Furthermore, in addition to demonstrating a notably strong alignment of our metric with radiologists' judgments on ReXVal~\cite{yu2023evaluating}, we introduce a new dataset containing tailored perturbations. 
These are designed to examine the sensitivity of evaluation metrics to significant versus insignificant modifications. 
Our findings indicate that our metric is sensitive to critical details such as pathologies or the location of findings, while remaining robust to errors of low significance, such as the omission of non-clinical phrases.

Hence, our contributions are three-fold:
\begin{enumerate}
    \item 
    A novel evaluation metric for medical report generation, considering both textual and clinical aspects.
    Its key idea is to measure  report similarity through their corresponding image.
    \item 
    Showing that our metric achieves the strongest alignment with radiologists judgment compared to previous metrics, as well as  the highest robustness to the perturbations. 
    \item 
    A new evaluation dataset containing targeted perturbations designed to reveal weaknesses and benefits in evaluation metrics for medical report generation.
\end{enumerate}

\section{Related Work}
\label{sec:related}

\noindent
{\bf Medical report generation.}
Medical report generation models aim to produce coherent and accurate medical reports based on input images. 
Most research in this field focuses on a paired setup, where both images and corresponding reports are available for training. 
Typically, visual features are encoded using a dedicated encoder module, followed by a decoder that generates the output report~\cite{chen-acl-2021-r2gencmn,chen2020generating,hou-etal-2023-organ,huang2023kiut,jing2017automatic,li2023unify,liu2019clinically,wang2022cross,zhang2020radiology}. 
Additional techniques, such as incorporating knowledge 
graphs~\cite{li2019knowledge,liu2021exploring,zhang2020radiology}, planning~\cite{hou-etal-2023-organ}, and reinforcement learning~\cite{gu2024complex,liu2019clinically,wang2022medical}, are often employed in the process. Training these models without access to paired data has also garnered attention in other works~\cite{hirsch2024medcycle,hirsch2024medrat,liu2021auto}. 

\noindent
{\bf Evaluation metrics for report generation.}
Evaluation is crucial for assessing the effectiveness of all the above approaches.
Traditional metrics used to measure alignment between two texts (reference and candidate) rely on n-gram overlap and include BLEU~\cite{papineni-etal-2002-bleu}, METEOR~\cite{banerjee-lavie-2005-meteor}, ROUGE~\cite{lin-2004-rouge} \& CIDEr~\cite{vedantam2015cider}.
These metrics consider exact word or phrase alignments, often overlooking paraphrases and failing to account for variations in semantic significance.
Specialized metrics such as clinical efficacy (CE) metrics have been developed to measure clinical precision, recall, F1 score, or cosine similarity. 
These metrics are typically derived from classification models trained to identify specific pathologies~\cite{irvin2019chexpert,smit2020combining}. 
While CE metrics assess high-level semantic similarity between reference and generated reports, they may not fully capture the coherence and overall quality of the generated reports.
Recently, another metric called RadGraph F1~\cite{yu2023evaluating} was proposed, which computes overlap between graphs representing clinical entities and relations in both reference and generated reports.
In~\cite{yu2023evaluating}, it is demonstrated that RadGraph, along with the embedding-based metric BERTScore~\cite{zhang2019bertscore}, exhibits relatively high alignment with radiologists' judgments.

\noindent
{\bf Image-report representation learning.}
Multi-modal representation learning in the medical domain is usually focused on Vision Language Pre-training (VLP), aiming to align corresponding images and texts in the same representation space. 
These representations can be utilized for various downstream tasks, such as retrieval, classification and detection. 
Most of the works in that domain utilize contrastive learning objectives to align the two modalities. 
ConVIRT~\cite{zhang2022contrastive} proposed a simple yet effective model that utilized a global contrastive objective between the image and the corresponding report. 
GLoRIA~\cite{huang2021gloria}, LIMITR~\cite{dawidowicz2023limitr} \& LoVT~\cite{muller2022joint} proposed different methods to utilize the local connections between the modalities and incorporated them into an additional contrastive objective.
MGCA~\cite{wang2022multi} added a prototype module which introduces disease-level information. 
Biovil~\cite{boecking2022making} focused on the language representations and introduced a domain-specific language model. 
MedCLIP~\cite{wang2022medclip} proposed to replace the contrastive loss with semantic matching loss based on medical knowledge to eliminate false negatives in contrastive learning.

\vspace{-0.1in}
\section{Proposed Metric}
\label{sec:proposed_metric}

The goal of report generation is to automatically create a free-text description for a clinical image. 
The quality of the generated reports should be quantified based on how well the overall information present in the image is expressed in the report. 
Currently, evaluation metrics measure the similarity between a single ground-truth report and the generated report without considering the images, focusing either on language similarity (NLG metrics) or clinical efficacy (CE metrics). 
NLG metrics do not account for the fact that certain sentences or specific words are more crucial than others, although these impact patient diagnosis and treatment more significantly. 
CE metrics compare pathologies but are limited in the types of pathologies they extract and ignore important pathology properties (e.g., location or severity). 
Additionally, this approach raises a question: how do we choose a model between two given ones when one is superior in terms of NLG and the other in terms of CE?

In the light of the limitations of existing metrics, we seek a new metric that quantifies the quality of a generated report, addressing the following criteria: (1) consideration of both textual and clinical aspects, (2)~recognition that subtle differences between reports may be significant, (3)~giving less weight to information that cannot be extracted from the image (e.g. previous conditions), and (4)~capturing multiple possible findings along with their properties such as locations and severity.

Our key idea is that the similarity between radiology reports should consider the corresponding radiology image. 
We propose to quantify the quality of the generated report with respect to both the information present in the input image and the ground-truth report. 
To integrate information from two modalities and compute a similarity score, we implement our idea at two steps. 
First, we project the image and reports into a shared space where similarities and differences are trained to be captured. 
Second, we propose a novel metric that quantifies quality based on the relationships among three components: the image, the generated report, and the ground-truth report, as described below.


\paragraph{Multi-modal embedding model.}
Medical image-report representation models aim to construct a joint embedding space where corresponding images and reports are positioned close to each other, while non-corresponding ones are placed farther apart~\cite{dawidowicz2023limitr,huang2021gloria,zhang2022contrastive,boecking2022making,wang2022medclip}. 
These models learn distinctive features that can be extracted from both the images and the reports. 
The properties of these embedding models align with desired properties in medical reports evaluation metrics:
 (1)~As the embedding model constructs the embedding space to successfully separate non-matching images and reports, it seeks the differences that matter, even when these are 
 subtle. 
 This is crucial for our purpose, as subtle differences between reports can distinguish between an excellent report and one that fails to describe the image. 
 Furthermore, the separation is learned rather than being guided by pre-defined criteria (e.g., specific $14$ pathologies). 
 Therefore, it enables the quantification of the quality of reports describing a wide range of pathologies and abnormalities.
 (2)~As the model learns to distinguish between examples within the dataset, it also acquires the ability to overlook general information lacking discriminative value.
 This property is important for an evaluation metric as it allows the metric to prioritize the extraction of pertinent clinical details from the input image.

Our approach can function with any shared space embedding model. 
However, we prioritize precise image-report retrieval-oriented models, i.e., finding the exact match from one modality given the other, as this aligns with our goal. 
Precise retrieval relies on matching fine details, which necessitates shared information, both local and global, within the image and report. 
Other focuses of embedding models are often on downstream tasks such as classification or segmentation, where the focus on specific pathologies usually suffices. 
Therefore, we chose to employ the model of~\cite{dawidowicz2023limitr}, which focuses on image-text and text-image retrieval, rather than on other downstream tasks. 
Its utilization aligns well with the task of evaluating report generation, which entails quantifying the accuracy of diverse descriptions contained within the report. 
Section~\ref{sec:analysis} demonstrates the model's benefit for our needs by showing its superior alignment with radiologists' judgment compared to other embedding models.

\paragraph{Report-report similarity evaluation.}
Given the embedding vectors of an image, a generated report, and a ground-truth report, our aim is to define a similarity score between them. 
There are various alternatives for such a score, as discussed in Section~\ref{subsec:ablation}. 

We  define the similarity as the area of the triangle created by the three high-dimensional points (i.e., the embedding vectors). 
Intuitively, this area takes into account the proximity to each other in the embedding space of all pairs of points in this triplet. 
Formally, for an image $i$, a generated report $g$, and a ground-truth report $r$, mapped by the embedding model $Emb$ to $i_e,g_e, r_e$ respectively, their proximity is computed using the following triangle area formula, utilizing the inner product $\langle \cdot, \cdot \rangle$:
\begin{equation}
    T(i, g, r) = \frac{1}{2} \sqrt{\langle i_e - g_e, i_e - g_e \rangle \langle i_e - r_e, i_e - r_e \rangle - \langle i_e - g_e, i_e - r_e \rangle ^2} \, 
\label{eq:dist}
\end{equation}
Since the embedding model is designed to position the image and the two reports that correspond to the same image close to each other in the embedding space, lower values of this measurement indicate a higher level of similarity between the reports and the image.

To normalize the measure to the range of $[0,1]$, where $0$ indicates low similarity and $1$ represents high similarity, the final quality score is defined as:
\begin{equation}
    VLScore(i, g, r) = \max \Bigl\{ 1 - \frac{T(i, g, r)}{C}, 0 \Bigr\} \; ,
\label{eq:VLS}
\end{equation}
where  $C$ is a constant.
We set it to $C=890$ to suit the maximal value of Equation~\ref{eq:dist} (representing the largest triangle area) for the baseline dataset~\cite{yu2023evaluating}, which comprises pairs of diverse quality. 
Therefore, this choice is compatible across a variety of datasets.

\section{Experimental Analysis}
\label{sec:analysis}

We analyze our proposed metric by comparing its alignment with radiologists' judgment and by using a controlled dataset to demonstrate the domain sensitivity of the various metrics across different error types. 
Further details on the experiments and findings are provided below.

\subsection{Radiologist-judgement data}
\label{sec:judgement}

This experiment aims to assess the agreement between evaluation metrics and the actual ratings of certified radiologists.
We expect good evaluation metrics to show strong correlation with the assessments of radiologists, who have extensive experience in interpreting medical reports and defining report accuracy.
A low correlation would raise concerns about the validity of the metric, since its scores would diverge from the opinions of radiologists, the experts in this field.

\noindent
\textbf{Dataset.}
The recent work of~\cite{yu2023evaluating} introduces an evaluation dataset, called {\em Radiology Report Expert Evaluation (ReXVal)}, comprising $200$ pairs of reports. 
Each pair consists of a reference report (drawn from the MIMIC-CXR training set) and a candidate report retrieved as the best match from the same dataset, according to one of the following automated metrics: BLEU, BERTScore, CheXbert, and RadGraph F1. 
Each pair was annotated by $6$ board-certified radiologists, who provide the number of errors in the candidate report across several error categories, such as omission of finding or false location of finding. 
These annotations generate a rating for each pair of reports based on the number of errors identified by the radiologists.

\noindent
\textbf{Correlation score.}
The alignment between the different metric scores (previous metrics \& ours) and the radiologist scores assigned to the same reports is quantified using the Kendall rank correlation coefficient ($\tau$), as proposed in~\cite{yu2023evaluating}. 
This coefficient assesses the degree of similarity in the ordering of ranks assigned by the metrics and radiologists.  
In this context, a high correlation coefficient indicates that the ranking produced by a given metric closely resembles the ranking assigned by radiologists. 
The agreement with radiologists is important to indicate the extent to which each metric captures the perceived quality or clinical relevance of the reports as assessed by experts.
This means the metric effectively identifies which report-reference pairs are considered better or worse. 

\noindent
\textbf{Results.}
Table~\ref{table:kendall} shows the correlation between common evaluation metrics and our proposed metric to the judgment of radiologists.
Our correlation coefficient ($\tau = 0.718$) is much higher than the classical evaluations (e.g., B-4 with $\tau=0.345$) and the recently proposed metric of RadGraph F1 (RG , $\tau = 0.515$).
This can be explained by radiologists' focus on the diagnostic significance of textual differences rather than on mere textual overlap. 
Our approach aims to achieve a similar focus, which contributes to the higher correlation we observe.

\begin{table}[t]
\caption{\textbf{Evaluation on ReXVal dataset.}
Compared to previous evaluation metrics, our metric achieves a higher correlation (Kendall coefficient $\tau$) with radiologist judgment from the ReXVal dataset.
We report the results of BLEU (B), METEOR (M), ROUGE-L (R-L), BERTScore (BS), CheXpert (Cp), CheXbert (Cb),  and RadGraph F1 (RG).}
\centering
\resizebox{1\linewidth}{!}{
\begin{tabular}{c|ccccc|ccc|c}
& \multicolumn{5}{c|}{NLG}& \multicolumn{3}{c|}{CE} &  OURS \\ 
 & B-1 & B-4 & M & R-L & BS & Cp& Cb & RG & VLScore \\ [0.01in] \hline
Kendall $\tau$ & 0.388 & 0.345 & 0.465 & 0.476 & 0.511 &  0.463& 0.499 &0.515 & \textbf{0.718} \\ \hline
\end{tabular}
}
\label{table:kendall}
\end{table}

\subsection{Perturbed data}

In the medical domain, the single dataset with human judgement is of~\cite{yu2023evaluating}, which we used in Section~\ref{sec:judgement} for metric assessment.
We propose to add another dateset, a {\em perturbed dataset,} which will assess the  robustness of the different metrics to several properties crucial for quantifying the accuracy of generated medical reports.
This dataset is of a larger scale and contains targeted perturbations that address several weaknesses that may arise in evaluation metrics and are present in generated reports.

Toward this end, we sampled a subset of $440$ pairs of images and reports from the validation and test sets of MIMIC-CXR. 
For this subset, we manually applied six types of modifications to the accompanying reports. 
This resulted in $979$ triplets, each includes an image, the ground-truth report and the perturbed report.
%

In order to quantify the quality of the metrics and assess for which modifications the metrics are robust or sensitive, we compare the behavior of each evaluation metric for each modification type. 
We discuss each pertubation and experiment below.


\noindent
\textbf{Removal of pathology sentence vs. insignificant sentence.} 
Different sentences in medical reports carry varying levels of significance. 
Those that describe the pathologies and diagnosis are central to the essence of the report, making their presence crucial in the generated report, while other sentences may be of lower significance.
To evaluate the sensitivity of the evaluation metrics to sentences with varying importance, we removed sentences from the ground-truth report, as follows.
In the first case, we examine the effect of a missing pathology by omitting the sentence that indicates it. 
Clinically, this difference is dramatic as it represents a major error in image analysis. 
Consequently, we expect the metrics to penalize considerably.
In this case, we obtained $238$ examples.

In the second case, we removed general sentences that are irrelevant to the analysis of the given image.
Such sentences could include common phrases that may or may not appear in reports (e.g., "a chest radiograph") or phrases related to the actions taken by the doctor (e.g., "called the nurse"). 
We do not expect a model to generate these sentences based on a given image. 
Therefore, we aim for the metric to be immune to the presence or absence of these phrases.
We obtained $119$ such examples.

Table~\ref{table:controlleddatasentence} shows the average similarity scores given by each metric for each one of these perturbation types
and their difference ($\Delta$).
A metric should penalize clinical errors more than general sentences.
However, both perturbations influence NLG metrics similarly (the difference in the similarity scores between the two is approximately 0), which shows their weak sensitivity to clinical details.
This behavior is expected as these metrics quantify the presence or absence of sequences of words without consideration of context and sentence importance. 
On the other hand, both CE metrics and our metric are aware of the clinical aspects and are affected differently by the two perturbations.
For CE metrics, which assess the overlap of clinical entities between two reports,  removing the pathology description greatly impact their similarity results.
Yet, our metric exhibits the highest sensitivity in this case (the highest $\Delta$).
It is derived from the nature of our metric relying on discriminative visual and textual information.
This is so since the embedding model utilizes clinical information from both the image and the report, mapping them to close or distant locations in a shared space. 
General information shared across multiple studies is assigned less weight in this mapping process, as it does not contribute significantly to the differentiation between radiology studies.

\begin{table}[t]
\caption{\textbf{Results on perturbed dataset: Removing pathology vs. insignificant sentences.}
When removing a pathology, the score should drop, while when omitting a general sentence, it should change only slightly. 
NLG metrics result in similar scores in both cases; 
however, CE and our metric penalize the score in accordance with the importance. 
Our metric shows the highest sensitivity.
}
\centering
\resizebox{1\linewidth}{!}{
\begin{tabular}{c|ccccc|ccc|c}
& \multicolumn{5}{c|}{NLG}& \multicolumn{3}{c|}{CE} &  OURS \\ 
Sentence Removal & B-1 & B-4 & M & R-L & BS & Cp& Cb & RG & VLScore  \\ [0.01in] \hline
 Insignificant   & 0.75 & 0.74 &0.49 & 0.85 & 0.89 & 0.99 & 0.97 & 0.93 & 0.84 \\
Significant (pathology) & 0.76 & 0.74 & 0.48 & 0.85 & 0.88 & 0.87 &0.90 & 0.82 & 0.69 \\ \hline
$\Delta$ & -0.01 & 0.00 & 0.01 & 0.00 & 0.01 & 0.12 & 0.07 & 0.11 & \textbf{0.15} \\ \hline
\end{tabular}
}
\label{table:controlleddatasentence}
\end{table}


\noindent
\textbf{Modification of pathology description word vs. non-informative word.} 
In medical reports, single words, often those that describe the pathology,
can significantly alter the diagnosis. 
To assess the sensitivity of the evaluation metrics to changes in such words, we modified a single word describing either the location or severity of the pathology. 
The modifications regarding the location of the pathology include alterations to other anatomical locations (e.g., basal to apical) and swaps between left and right. 
As for the severity of the pathology, we modified the severity level. 
To establish a baseline for this evaluation, we also replaced non-informative words (e.g., "the" / "this" / "there") by the [UNK] token. 
We anticipate that a change in the pathology description will exert a greater influence on the metrics than a change in non-informative words.
We obtained $187$ examples for the location change, $95$ for the severity change and $133$ for the non-informative word change.

Table~\ref{table:controlleddataword} presents the difference ($\Delta$) of each metric for the significant perturbations (location or severity) relative to the non-informative word change.
As before, NLG metrics are unaware of the significance of the words. 
The scores are similar for both significant and non-significant word perturbations.
Additionally, for all these perturbations, most of the scores (all but METEOR) are very close to $1$, suggesting almost perfect similarity, although the clinical difference caused by the perturbations (e.g., mislocating a finding) are very significant. 
CE metrics also fail to recognize the significance of these perturbations. 
The difference to the non-informative word perturbations for these metrics is low, and the absolute scores are relatively high (close to $1$), demonstrating the low sensitivity of these metrics to such errors. 
This is explained by the fact that these metrics often focus on the existence of findings but not their details. 
Our metric, however, is aware of these subtle differences and their importance, due to the contrastive-based design of the shared space we operate within, leading to a  penalty in the score compared to the change of a non-informative word.

\begin{table}[t]
\caption{\textbf{Results on perturbed dataset: Modifying a descriptive vs. non-informative word. }
When altering a clinical descriptor, the score should decrease, while when changing a non-informative word, it should change only slightly. 
NLG \& CE metrics yield similar scores in these cases; 
Our metric shows high sensitivity.
Moreover, all previous metrics set a high score (close to $1$) to the perturbed report, suggesting it is near perfect, although a critical error was made.
}
\centering
\resizebox{1\linewidth}{!}{
\begin{tabular}{c|ccccc|ccc|c}
& \multicolumn{5}{c|}{NLG}& \multicolumn{3}{c|}{CE} &  OURS \\ 
Changed Word & B-1 & B-4 & M & R-L & BS & Cp & Cb & RG & VLScore \\ [0.01in] \hline
Non-informative & 0.98 & 0.95 & 0.65 & 0.98 & 0.97 & 0.99 & 0.99 & 0.97 & 0.91 \\ \hline \hline
Location & 0.98 & 0.95 & 0.67 & 0.98 & 0.99 & 1.0 & 1.0 & 0.93 & 0.72 \\
$\Delta$ & 0.00 & 0.00 & -0.02 & 0.00 & -0.02 & -0.01 & -0.01 & 0.04 & \textbf{0.19} \\ \hline
Severity & 0.97 & 0.94 & 0.65 & 0.97 & 0.99 & 0.98 & 0.99 & 0.92 & 0.79 \\
$\Delta$ & 0.01 & 0.01 &0.00  &0.01  &  -0.02& 0.01 & 0.00  & 0.05 & \textbf{0.12} \\ \hline
\end{tabular}
}
\label{table:controlleddataword}
\end{table}

\begin{figure}[t]
\centering
\begin{tabular}{cccc}
     \includegraphics[width=0.22\linewidth]{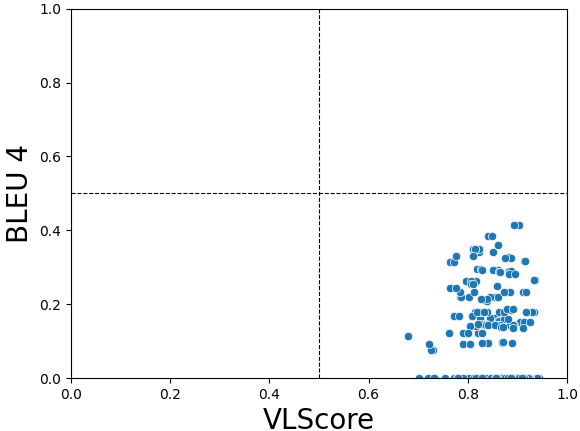} &
     \includegraphics[width=0.22\linewidth]{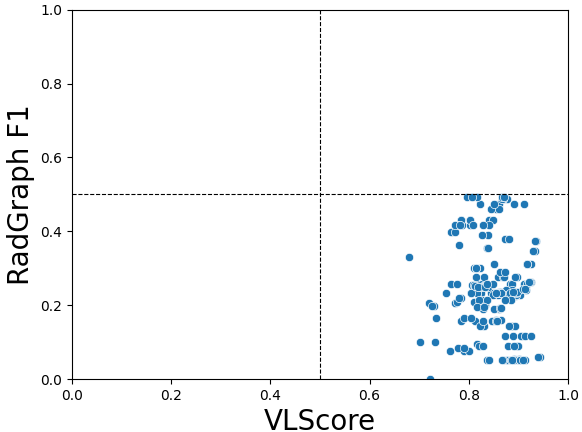} &
     \includegraphics[width=0.22\linewidth]{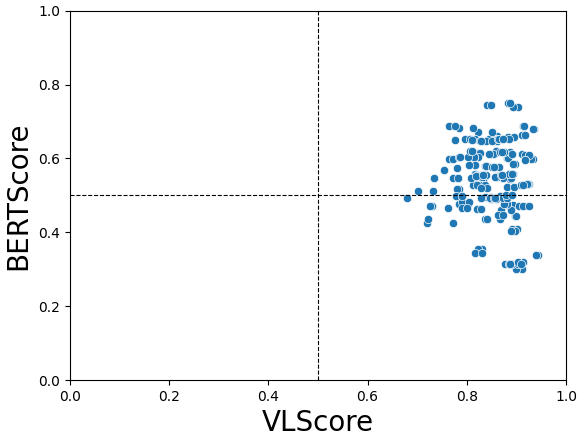} &
     \includegraphics[width=0.22\linewidth]{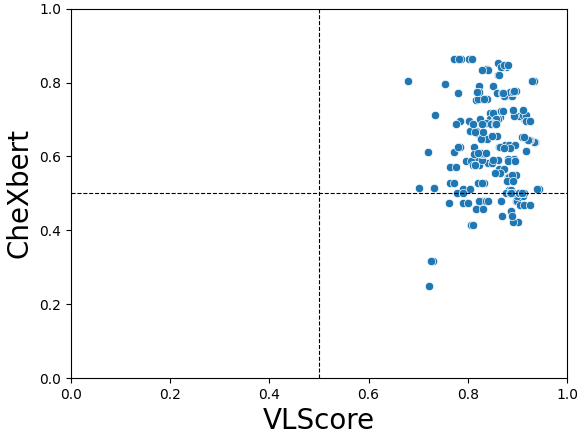}
     \\
    (a) Ours vs. B-4 & (b) Ours vs. RG & (c) Ours vs. BS & (d) Ours vs. Cb
\end{tabular}
\caption{{\bf Scores of equivalent normal reports.} 
When comparing two reports with no findings, the score is expected to be high, as both convey the same clinical findings with differences only in the writing of the report.
Compared to BLEU-4 (a) and RadGraph F1 (b), our metric yields high scores (right-side), while the other metrics yield low scores (bottom-side). 
BERTScore (c) provides mid-range scores instead of high scores.
CheXpert (d) provides higher scores than the others, yet lower scores than our metric, although it was expected to yield scores close to $1$ as both reports in each pair contain no findings.
}
\label{fig:nofindings}
\end{figure}

\noindent
\textbf{Modification of reports  without findings.}
There is no single structure for medical reports. 
Consequently, a radiology image can be described in various ways depending on factors such as the patient's history or the radiologist's style and focus. 
However, the core purpose of the report -- diagnosing abnormalities and recognizing normal findings, whether explicitly or implicitly -- should remain consistent.
To demonstrate the robustness of metrics to these variations, 
we specifically select studies with normal findings only.
We created a standardized general report representing studies with no findings by compiling the most common sentences from the dataset associated with reports lacking abnormalities. 
Then, for each image corresponding to a report with no abnormalities, we pair it with this constructed report (see example in Figure~\ref{fig:teaser}a). 
In total, we obtain $207$ such examples.
In this experiment, we expect a metric to exhibit robustness to the exact wording and to discern the clinical meaning of the report. 
Specifically, the metric score should be relatively high.

Figure~\ref{fig:nofindings} compares the scores of our metric to those given by  four other metrics.
BLEU-4 and RadGraph F1 assign these examples a low score, particularly for BLEU-4 the average is $0.15$ and for RadGraph F1 is $0.23$, whereas BERTScore yields mid-level results with an average of $0.51$. 
This occurs because BLEU and RadGraph are sensitive to the exact wording, while BERTScore exhibits some robustness to words with similar meanings.
CheXbert, which is a CE metric, is expected to assign very high scores in this case, as all the reports should be classified with no findings.
Yet,  its average is only $0.63$.
The other CE metric, CheXpert (F1), which directly measures classification accuracy, does behave as expected with an average of $0.91$.
Our scores are predominantly high, with an average of $0.85$, comparable to CheXpert and surpass all other methods, as we do not rely solely on text similarity but rather on information that is mutual to the visual and textual modalities, with semantics captured in the shared space. 
Other metrics demonstrate undesirable behavior as those depicted in Figure~\ref{fig:nofindings}: BLEU-1 $0.33$, METEOR $0.18$, ROUGE-L $0.33$.

\subsection{Ablation study}
\label{subsec:ablation}

\noindent
\textbf{Similarity measurements.}
Given three embedding vectors--one for the ground-truth report $r_e$, one for the generated report $g_e$, and one for the input image $i_e$---Table~\ref{table:similarity measurmenets} presents the results of several alternative similarity measurements.
For all the measures, we examine the correlation with the ranking of radiologists, using the Kendall correlation coefficient. 
The first measure considers only $r_e$ and $g_e$, and the others consider also the image embedding $i_e$.
These measures are:
(1)~The cosine similarity between $r_e$ and $g_e$.
This distance implicitly considers the image as part of the construction of the shared space, but not in the computation of the metric. 
(2)~The radius of the minimal bounding sphere measures the density of the three points in the high dimensional space by estimating the minimal bounding sphere of these three points.
(3)~Image centered cosine similarity is the cosine similarity between the difference vectors $r_e$-$i_e$ and $g_e$-$i_e$.
The difference between this similarity and the vanilla cosine similarity is the origin, i.e. whether it is the $\vec{0}$ vector or the image vector.
(4)~Our metric: the area of the triangle created by the three high-dimensional points (embedding vectors).

Table~\ref{table:similarity measurmenets} shows that the results by these measurements, which are all based explicitly or implicitly on connections with visual information, are all superior to the previous methods presented in Table~\ref{table:kendall}, which are purely textual.
Furthermore, the relative position of the image in the embedding space to the two reports is useful for the similarity measurement, as evident by the higher correlation achieved in the image-centered cosine similarity compared to the regular cosine similarity ($0.678$ vs. $0.555$).
In addition, when considering the Euclidean distances between the points and not only the angles between them, as captured by the triangle area, the correlation boosts ($0.718$).
The minimal bounding sphere shows relatively low correlation ($0.525$), possibly due to the exponential growth of its volume in high dimensions.
Intuitively,
bounding spheres might contain mostly empty volume.

\begin{table}[t]
\caption{\textbf{Similarity measurements ablation.}  Our proposed metric, based on triangle area computation, achieves the highest correlation with radiologists' ranking (Kendall's $\tau$ coefficient on ReXVal dataset), compared to other methods to measure the similarity of embeddings in the shared space.
}
\centering
\begin{tabular}{c|c|ccc}
\multirow{2}{*}
&  Cosine similarity & \makecell{Minimal bounding \\ sphere} & \makecell{Image-centered \\ cosine similarity}  & \makecell{Triangle area \\ (ours)} \\ \hline 
Kendall $\tau$ & 0.555 &0.525 & 0.678 & \textbf{0.718}  \\ \hline
\end{tabular}
\label{table:similarity measurmenets}
\end{table}

\noindent
\textbf{Embedding models.}
Recall that we used LIMITR~\cite{dawidowicz2023limitr} as our multi-modal embedding model due to its focus on precise retrieval tasks, as well as extracting discriminative representations at both local and global levels. 
Table~\ref{table:embedding models} confirms that this model outperforms other multi-modal embedding models.
LIMITR embeddings demonstrate the highest correlation with radiologists' rankings compared to ConVIRT~\cite{zhang2022contrastive}, BioViL~\cite{boecking2022making}, MedCLIP~\cite{wang2022medclip}, and GLoRIA~\cite{huang2021gloria} (the $C$ values set for these models are $124$, $1$, $0.06$, and $1155$, respectively).
  This may be explained by the fact that LIMITR is optimized for retrieval tasks, excelling at matching a single image with a specific report, while the other models prioritize downstream tasks such as classification and segmentation, which may be less precise for capturing nuances relevant for evaluation purposes.
 Since our metric relies on strong alignment between the image and the entire report to assess the quality of generated reports, the emphasis of LIMITR on alignment-focused tasks aligns well with our evaluation criteria.

\begin{table}[t]
\caption{\textbf{Embedding models ablation.} Comparison of various vision-language embedding models to construct the shared space for our metric. 
LIMITR achieves the highest correlation with radiologists judgment (Kendall’s $\tau$ coefficant on ReXVal dataset).
}
\centering
\begin{tabular}{c|ccccc}
&   ConVIRT~\cite{zhang2022contrastive} & BioViL~\cite{boecking2022making} & MedCLIP~\cite{wang2022medclip} & GLoRIA~\cite{huang2021gloria} & LIMITR~\cite{dawidowicz2023limitr}  \\ [0.01in] \hline
Kendall $\tau$ &0.526& 0.580 &0.636 & 0.698 & \textbf{0.718}  \\ \hline
\end{tabular}
\label{table:embedding models}
\end{table}


{\bf Limitations.}
The constant $C$ utilized in Eq.~\ref{eq:VLS} is determined based on a representative dataset and is applicable to all datasets within the domain. 
It serves the purpose of normalizing the size of the triangle to the range $[0,1]$ (if desired), as the size of the triangle is unbounded. 
In different domains, such as other imaging sources, the value of $C$ will need to be determined accordingly.

{\bf Societal impact.} 
The high correlation of our metric with radiologists' judgment may result in overreliance on automated systems, which are prone to errors. 
If a metric leads to increased reliance on automated systems for generating medical reports, there is a risk that healthcare professionals may become overly dependent on these technologies. 
This could potentially lead to a reduction in critical thinking and clinical judgment, as well as a loss of the human touch in patient care. 
Therefore, while this metric can be used to quantify the quality of automated systems, their utilization in real-world practice should be carefully supervised.

\vspace{-0.05in}
\section{Conclusions}
This paper introduces a novel evaluation metric, $VLScore$, for automatic medical report generation from X-ray images. 
It addresses shortcomings found in current evaluation methods, which either focus solely on textual similarities, neglecting clinical considerations, or concentrate solely on one clinical aspect, such as pathology, while disregarding all other facets. 
The main idea behind our metric is to assess the similarity between radiology reports in relation to their corresponding radiology image. 
The metric measures similarity in a joint visual-textual space, according to the triplet of embeddings, of the image, the ground-truth report, and the generated report. 
We demonstrate the strength of our metric compared to previous metrics both qualitatively and quantitatively; for instance, it improves the correlation with radiologists' judgments by $20\%$ on the ReXVal dataset. 
In addition, we propose a new dataset for metric evaluation, featuring thoughtfully crafted perturbations that differentiate between substantial modifications (e.g., omitting a diagnosis) and less significant ones.



{\small
\bibliographystyle{splncs04}
\bibliography{egbib}

\begin{thebibliography}{10}
\providecommand{\url}[1]{\texttt{#1}}
\providecommand{\urlprefix}{URL }
\providecommand{\doi}[1]{https://doi.org/#1}

\bibitem{banerjee-lavie-2005-meteor}
Banerjee, S., Lavie, A.: {METEOR}: An automatic metric for {MT} evaluation with improved correlation with human judgments. In: Proceedings of the {ACL} Workshop on Intrinsic and Extrinsic Evaluation Measures for Machine Translation and/or Summarization (2005)

\bibitem{boecking2022making}
Boecking, B., Usuyama, N., Bannur, S., Castro, D.C., Schwaighofer, A., Hyland, S., Wetscherek, M., Naumann, T., Nori, A., Alvarez-Valle, J., et~al.: Making the most of text semantics to improve biomedical vision--language processing. In: European conference on computer vision. pp. 1--21. Springer (2022)

\bibitem{chen-acl-2021-r2gencmn}
Chen, Z., Shen, Y., Song, Y., Wan, X.: Cross-modal memory networks for radiology report generation. In: Proceedings of the 59th Annual Meeting of the Association for Computational Linguistics and the 11th International Joint Conference on Natural Language Processing. Association for Computational Linguistics (2021)

\bibitem{chen-etal-2020-generating}
Chen, Z., Song, Y., Chang, T.H., Wan, X.: Generating radiology reports via memory-driven transformer. In: Proceedings of the 2020 Conference on Empirical Methods in Natural Language Processing (EMNLP). Association for Computational Linguistics (2020)

\bibitem{chen2020generating}
Chen, Z., Song, Y., Chang, T.H., Wan, X.: Generating radiology reports via memory-driven transformer. In: Proceedings of the 2020 Conference on Empirical Methods in Natural Language Processing (EMNLP). pp. 1439--1449 (2020)

\bibitem{dawidowicz2023limitr}
Dawidowicz, G., Hirsch, E., Tal, A.: Limitr: Leveraging local information for medical image-text representation. In: Proceedings of the IEEE/CVF International Conference on Computer Vision. pp. 21165--21173 (2023)

\bibitem{gu2024complex}
Gu, T., Liu, D., Li, Z., Cai, W.: Complex organ mask guided radiology report generation. In: Proceedings of the IEEE/CVF Winter Conference on Applications of Computer Vision. pp. 7995--8004 (2024)

\bibitem{hirsch2024medcycle}
Hirsch, E., Dawidowicz, G., Tal, A.: Medcycle: Unpaired medical report generation via cycle-consistency. In: Findings of the Association for Computational Linguistics: NAACL 2024. pp. 1929--1944 (2024)

\bibitem{hirsch2024medrat}
Hirsch, E., Dawidowicz, G., Tal, A.: Medrat: Unpaired medical report generation via auxiliary tasks. arXiv preprint arXiv:2407.03919  (2024)

\bibitem{hou-etal-2023-organ}
Hou, W., Xu, K., Cheng, Y., Li, W., Liu, J.: {ORGAN}: Observation-guided radiology report generation via tree reasoning. In: Rogers, A., Boyd-Graber, J., Okazaki, N. (eds.) Proceedings of the 61st Annual Meeting of the Association for Computational Linguistics (Volume 1: Long Papers). pp. 8108--8122. Association for Computational Linguistics, Toronto, Canada (Jul 2023)

\bibitem{huang2021gloria}
Huang, S.C., Shen, L., Lungren, M.P., Yeung, S.: Gloria: A multimodal global-local representation learning framework for label-efficient medical image recognition. In: Proceedings of the IEEE/CVF International Conference on Computer Vision. pp. 3942--3951 (2021)

\bibitem{huang2023kiut}
Huang, Z., Zhang, X., Zhang, S.: Kiut: Knowledge-injected u-transformer for radiology report generation. In: Proceedings of the IEEE/CVF Conference on Computer Vision and Pattern Recognition. pp. 19809--19818 (2023)

\bibitem{irvin2019chexpert}
Irvin, J., Rajpurkar, P., Ko, M., Yu, Y., Ciurea-Ilcus, S., Chute, C., Marklund, H., Haghgoo, B., Ball, R., Shpanskaya, K., et~al.: Chexpert: A large chest radiograph dataset with uncertainty labels and expert comparison. In: Proceedings of the AAAI conference on artificial intelligence. vol.~33, pp. 590--597 (2019)

\bibitem{jain2021radgraph}
Jain, S., Agrawal, A., Saporta, A., Truong, S.Q., Duong, D.N., Bui, T., Chambon, P., Zhang, Y., Lungren, M.P., Ng, A.Y., et~al.: Radgraph: Extracting clinical entities and relations from radiology reports. arXiv preprint arXiv:2106.14463  (2021)

\bibitem{jing2017automatic}
Jing, B., Xie, P., Xing, E.: On the automatic generation of medical imaging reports. arXiv preprint arXiv:1711.08195  (2017)

\bibitem{johnson2019mimic}
Johnson, A.E., Pollard, T.J., Berkowitz, S.J., Greenbaum, N.R., Lungren, M.P., Deng, C.y., Mark, R.G., Horng, S.: Mimic-cxr, a de-identified publicly available database of chest radiographs with free-text reports. Scientific data  \textbf{6}(1), ~317 (2019)

\bibitem{li2019knowledge}
Li, C.Y., Liang, X., Hu, Z., Xing, E.P.: Knowledge-driven encode, retrieve, paraphrase for medical image report generation. In: Proceedings of the AAAI Conference on Artificial Intelligence. vol.~33, pp. 6666--6673 (2019)

\bibitem{li2023dynamic}
Li, M., Lin, B., Chen, Z., Lin, H., Liang, X., Chang, X.: Dynamic graph enhanced contrastive learning for chest x-ray report generation. In: Proceedings of the IEEE/CVF Conference on Computer Vision and Pattern Recognition. pp. 3334--3343 (2023)

\bibitem{li2023unify}
Li, Y., Yang, B., Cheng, X., Zhu, Z., Li, H., Zou, Y.: Unify, align and refine: Multi-level semantic alignment for radiology report generation. In: Proceedings of the IEEE/CVF International Conference on Computer Vision. pp. 2863--2874 (2023)

\bibitem{lin-2004-rouge}
Lin, C.Y.: {ROUGE}: A package for automatic evaluation of summaries. In: Text Summarization Branches Out. Association for Computational Linguistics (Jul 2004)

\bibitem{liu2021exploring}
Liu, F., Wu, X., Ge, S., Fan, W., Zou, Y.: Exploring and distilling posterior and prior knowledge for radiology report generation. In: Proceedings of the IEEE/CVF conference on computer vision and pattern recognition. pp. 13753--13762 (2021)

\bibitem{liu2021auto}
Liu, F., You, C., Wu, X., Ge, S., Sun, X., et~al.: Auto-encoding knowledge graph for unsupervised medical report generation. Advances in Neural Information Processing Systems  \textbf{34},  16266--16279 (2021)

\bibitem{liu2019clinically}
Liu, G., Hsu, T.M.H., McDermott, M., Boag, W., Weng, W.H., Szolovits, P., Ghassemi, M.: Clinically accurate chest x-ray report generation. In: Machine Learning for Healthcare Conference. pp. 249--269. PMLR (2019)

\bibitem{muller2022joint}
M{\"u}ller, P., Kaissis, G., Zou, C., Rueckert, D.: Joint learning of localized representations from medical images and reports. In: European Conference on Computer Vision. pp. 685--701. Springer (2022)

\bibitem{papineni-etal-2002-bleu}
Papineni, K., Roukos, S., Ward, T., Zhu, W.J.: {B}leu: a method for automatic evaluation of machine translation. In: Proceedings of the 40th Annual Meeting of the Association for Computational Linguistics (2002)

\bibitem{smit2020combining}
Smit, A., Jain, S., Rajpurkar, P., Pareek, A., Ng, A.Y., Lungren, M.: Combining automatic labelers and expert annotations for accurate radiology report labeling using bert. In: Proceedings of the 2020 Conference on Empirical Methods in Natural Language Processing (EMNLP). pp. 1500--1519 (2020)

\bibitem{tanida2023interactive}
Tanida, T., M{\"u}ller, P., Kaissis, G., Rueckert, D.: Interactive and explainable region-guided radiology report generation. In: Proceedings of the IEEE/CVF Conference on Computer Vision and Pattern Recognition. pp. 7433--7442 (2023)

\bibitem{vedantam2015cider}
Vedantam, R., Lawrence~Zitnick, C., Parikh, D.: Cider: Consensus-based image description evaluation. In: Proceedings of the IEEE conference on computer vision and pattern recognition. pp. 4566--4575 (2015)

\bibitem{wang2022multi}
Wang, F., Zhou, Y., Wang, S., Vardhanabhuti, V., Yu, L.: Multi-granularity cross-modal alignment for generalized medical visual representation learning. Advances in Neural Information Processing Systems  \textbf{35},  33536--33549 (2022)

\bibitem{wang2022cross}
Wang, J., Bhalerao, A., He, Y.: Cross-modal prototype driven network for radiology report generation. In: Computer Vision--ECCV 2022: 17th European Conference, Tel Aviv, Israel, October 23--27, 2022, Proceedings, Part XXXV. pp. 563--579. Springer (2022)

\bibitem{wang2023metransformer}
Wang, Z., Liu, L., Wang, L., Zhou, L.: Metransformer: Radiology report generation by transformer with multiple learnable expert tokens. In: Proceedings of the IEEE/CVF Conference on Computer Vision and Pattern Recognition. pp. 11558--11567 (2023)

\bibitem{wang2022medical}
Wang, Z., Tang, M., Wang, L., Li, X., Zhou, L.: A medical semantic-assisted transformer for radiographic report generation. In: Medical Image Computing and Computer Assisted Intervention--MICCAI 2022. pp. 655--664. Springer (2022)

\bibitem{wang2022medclip}
Wang, Z., Wu, Z., Agarwal, D., Sun, J.: Medclip: Contrastive learning from unpaired medical images and text (2022)

\bibitem{yu2023evaluating}
Yu, F., Endo, M., Krishnan, R., Pan, I., Tsai, A., Reis, E.P., Fonseca, E.K.U.N., Lee, H.M.H., Abad, Z.S.H., Ng, A.Y., et~al.: Evaluating progress in automatic chest x-ray radiology report generation. Patterns  \textbf{4}(9) (2023)

\bibitem{zhang2019bertscore}
Zhang, T., Kishore, V., Wu, F., Weinberger, K.Q., Artzi, Y.: Bertscore: Evaluating text generation with bert. In: International Conference on Learning Representations (2019)

\bibitem{zhang2020radiology}
Zhang, Y., Wang, X., Xu, Z., Yu, Q., Yuille, A., Xu, D.: When radiology report generation meets knowledge graph. In: Proceedings of the AAAI Conference on Artificial Intelligence. vol.~34, pp. 12910--12917 (2020)

\bibitem{zhang2022contrastive}
Zhang, Y., Jiang, H., Miura, Y., Manning, C.D., Langlotz, C.P.: Contrastive learning of medical visual representations from paired images and text. In: Machine Learning for Healthcare Conference. pp. 2--25. PMLR (2022)

\end{thebibliography}
}

\end{document}